\documentclass[final]{cvpr}

\usepackage{times}
\usepackage{epsfig}
\usepackage{graphicx}
\usepackage{amsmath}
\usepackage{amssymb}
\usepackage{makecell}
\usepackage{algorithm, algorithmic}
\usepackage{amsmath}
\usepackage{eqparbox}
\usepackage{subfigure}
\usepackage{mathrsfs}
\usepackage{url}
\usepackage{multirow}
\usepackage{appendix}

\usepackage[pagebackref=true,breaklinks=true,colorlinks,bookmarks=false]{hyperref}

\newcommand{\x}{\boldsymbol{x}}
\newcommand{\B}{\textbf{B}}
\newcommand{\z}{\boldsymbol{z}}

\begin{document}

\title{Transferable Semantic Augmentation for Domain Adaptation}

\author{
 Shuang Li\textsuperscript{1} \space
 Mixue Xie\textsuperscript{1} \space
 Kaixiong Gong\textsuperscript{1} \space
 Chi Harold Liu\textsuperscript{1\thanks{Corresponding author.}} \space\space
 Yulin Wang\textsuperscript{\rm2} \space
 Wei Li\textsuperscript{3} \space \vspace{.3em}\\
 \textsuperscript{1}Beijing Institute of Technology \quad \textsuperscript{2}Tsinghua University\quad \textsuperscript{3}Inceptio Tech. \\
 {\tt\small shuangli@bit.edu.cn \space michellexie102@gmail.com \space kxgong@bit.edu.cn \space liuchi02@gmail.com} \\
 {\tt\small wang-yl19@mails.tsinghua.edu.cn \space liweimcc@gmail.com}\\
}

\maketitle
\pagestyle{empty}
\thispagestyle{empty}

\begin{abstract}
   Domain adaptation has been widely explored by transferring the knowledge from a label-rich source domain to a related but unlabeled target domain. Most existing domain adaptation algorithms attend to adapting feature representations across two domains with the guidance of a shared source-supervised classifier. However, such classifier limits the generalization ability towards unlabeled target recognition. To remedy this, we propose a Transferable Semantic Augmentation (TSA) approach to enhance the classifier adaptation ability through implicitly generating source features towards target semantics. Specifically, TSA is inspired by the fact that deep feature transformation towards a certain direction can be represented as meaningful semantic altering in the original input space. Thus, source features can be augmented to effectively equip with target semantics to train a more transferable classifier. To achieve this, for each class, we first use the inter-domain feature mean difference and target intra-class feature covariance to construct a multivariate normal distribution. Then we augment source features with random directions sampled from the distribution class-wisely. Interestingly, such source augmentation is implicitly implemented through an expected transferable cross-entropy loss over the augmented source distribution, where an upper bound of the expected loss is derived and minimized, introducing negligible computational overhead. As a light-weight and general technique, TSA can be easily plugged into various domain adaptation methods, bringing remarkable improvements. Comprehensive experiments on cross-domain benchmarks validate the efficacy of TSA.
\end{abstract}

\section{Introduction}

\begin{figure}[htbp]\centering
    \setlength{\abovecaptionskip}{0.cm}
    \setlength{\belowcaptionskip}{0.cm}
    \centering
    \includegraphics[width=0.465\textwidth]{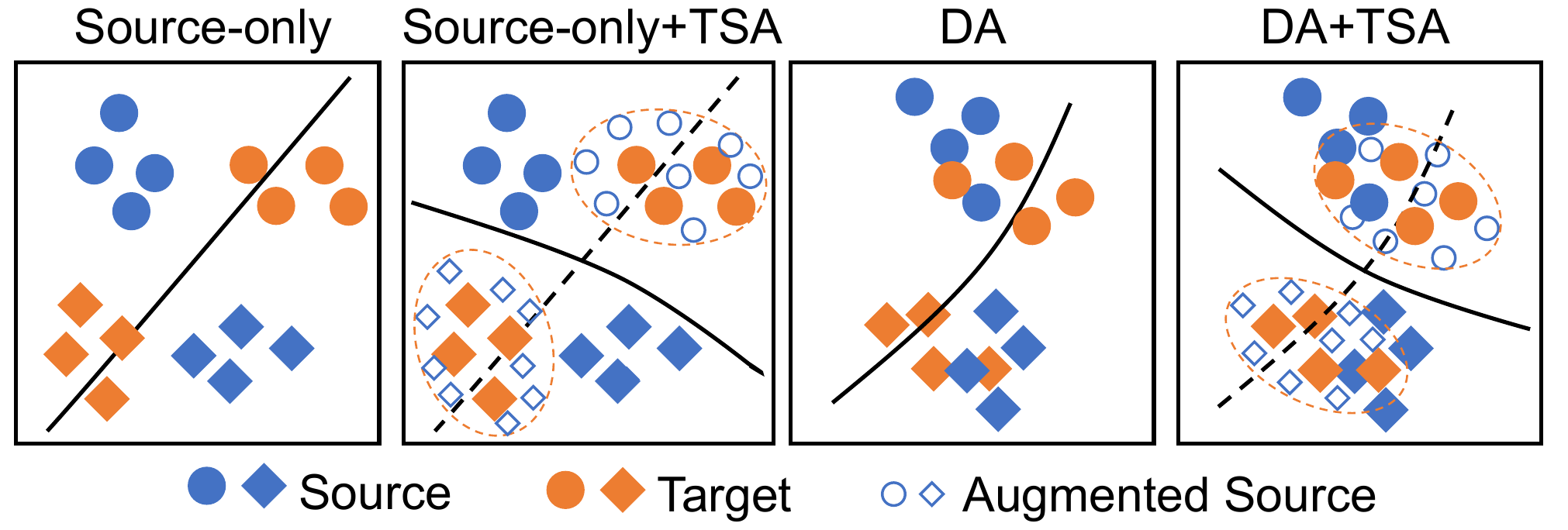}
    \caption{Overview of TSA, which relaxes the assumption that source and target domains share the same classifier. For both traditional deep learning and DA methods, TSA could augment source features towards target semantics to successfully adapt the final classifier from source domain to target domain.}
    \label{Fig_intro}
    \vspace{-3mm}
\end{figure}

Deep learning has achieved remarkable success on various vision tasks, including image recognition \cite{alexnet,resnet,densenet} and semantic segmentation \cite{FDNet,SEAN}. However, the recent success of deep learning methods heavily relies on massive labeled data. In practice, collecting abundant annotated data is expensive \cite{survey, DA-survey, Review_of_SingleSourceUDA, survey_of_VDA}. Meanwhile, each domain has its own specific exploratory factors, namely semantics, e.g., the illuminations, colors, visual angles or backgrounds, resulting in the domain shift \cite{DRMEA}. Hence, traditional deep models trained on a large dataset usually show poor generalizations on a new domain due to the domain shift issues \cite{DAN, DANN}. To remedy this, one appealing alternative is domain adaptation (DA), which strives to leverage the knowledge of a label-rich source domain to assist the learning in a related but unlabeled target domain \cite{DAN, DANN}.

Prior deep DA methods can be roughly categorized as 1) statistical discrepancy minimization based methods \cite{JAN,DRCN,MDD}, which leverage statistical regularizations to explicitly mitigate the cross-domain distribution discrepancy; and 2) adversarial learning based methods \cite{JADA,CDAN,DANN}, which strive to learn domain-invariant representations across two domains via adversarial manners.

Indeed, these DA methods have admittedly yielded promising results, but most of them assume a shared classifier with domain-invariant representations derived. Rare attention has been paid to explicitly enhancing the adaptation ability of the source-supervised classifier, which is also fundamental to DA problems as shown in Fig. \ref{Fig_intro}. To achieve classifier adaptation, Long et al. \cite{RTN} introduce classifier residual learning to explicitly model the classifier difference across domains. SymNets \cite{SymNets} constructs three classifiers to facilitate joint distribution alignment. However, they all rely on designing complex network architectures, which may suffer from high computational overhead and hinder the capability and versatility of these methods.

To alleviate aforementioned issues, we propose a Transferable Semantic Augmentation (TSA) approach to implicitly augment source features with target semantic guidance in the deep feature space to facilitate classifier adaptation. Specifically, TSA is motivated by the intriguing property that deep networks excel at disentangling the underlying factors of data variation and linearizing the deep features \cite{DFI, ISDA}. There exist many different semantic transformation directions in the deep feature space, and the semantic transformation of one sample can be enforced by translating its deep feature along a certain direction, such as the direction of changing backgrounds. However, it is nontrivial to explicitly discover all kinds of semantic transformation directions. In addition, not all directions are meaningful.

Hence, to effectively explore meaningful transformation directions, we first estimate the inter-domain feature mean difference for each class as the class-wise overall semantic (i.e., an integration of various semantics in one class) bias in the deep feature space. Besides, since the specific semantic information (e.g., different backgrounds, shapes or visual angles) of source and target are different, TSA further estimates target intra-class feature covariance to effectively capture the intra-class semantic variations of target domain. To obtain the accurate estimation, we introduce a memory module to class-wisely calculate feature mean and covariance with pseudo-labeled target samples. At last, we sample semantic transformation directions for source augmentation from a multivariate normal distribution, with the estimated feature mean difference as mean and the target intra-class covariance. In this way, the overall semantic difference between domains and the target intra-class semantic variations can guide source augmented features towards target.

Furthermore, to avoid explicitly generating augmented features and improve the efficiency of TSA, we develop an expected transferable cross-entropy loss over the augmented source distribution with an upper bound derived. By minimizing the upper bound of the expected loss, the source semantic augmentation can be performed and the extra computational overhead is negligible. Then, the trained source classifier can be successfully adapted to target. 

Contributions of this work are summarized as follows: 
\begin{itemize}
    \item We propose a novel Transferable Semantic Augmentation (TSA) method for classifier adaptation, which enables source feature augmentation towards target in an implicit manner. Notably, TSA introduces no extra network modules over the backbone network, making it simple to implement and computationally efficient.
    \item We develop a novel expected transferable cross-entropy loss over the augmented source distribution for DA, which greatly enhances the classifier adaptation ability. Moreover, as a light-weight and general technique, TSA can be easily plugged into various DA methods to significantly boost their performances. 
    \item Extensive experiments on several cross-domain benchmarks, including Office-Home, Office-31, VisDA-2017 and digits demonstrate that TSA can consistently yield significant performance improvements.
\end{itemize}

\section{Related Work}

\subsection{Feature Adaptation}

Existing domain-invariant feature learning methods mainly fall into two categories. One is the discrepancy-based works \cite{DAN, JAN, DRCN, MDD, DRMEA, TPN, BNM, ETD}, which concentrates on mitigating the domain difference by minimizing some statistical discrepancy metrics. To name a few, DAN \cite{DAN} minimizes maximum mean discrepancy (MMD) on task-specific layers to close the domain gap. Further, JAN in \cite{JAN} introduces a joint MMD to enforce joint distribution alignment between domains. \cite{MDD} proposes margin disparity discrepancy (MDD) to facilitate precise alignment with theoretical guarantees. Based on the optimal transport (OT) distance \cite{OT}, \cite{ETD} proposes an enhanced transport distance (ETD) to learn the optimal transport plan.

Another category is adversarial-based works \cite{MCD, TADA, DANN, CDAN, JADA}, which learns domain-invariant features in adversarial manners. For example, DANN \cite{DANN} introduces a domain discriminator to force the features of two domains indistinguishable. Following DANN, CDAN \cite{CDAN} integrates classification information into the domain discriminator to alleviate class-wise mismatching. While MCD \cite{MCD} employs a new adversarial paradigm where the adversarial manner occurs between the feature extractor and classifiers rather than the feature extractor and the domain discriminator.

In contrast, instead of learning domain-invariant features, TSA harnesses the augmented source features to ameliorate the generalization ability of source classifier, which is also crucial for tackling DA problems.

\begin{figure*}[htbp]\centering
    \setlength{\abovecaptionskip}{0.cm}
    \setlength{\belowcaptionskip}{0.cm}
    \centering
    \includegraphics[width=1.0\textwidth]{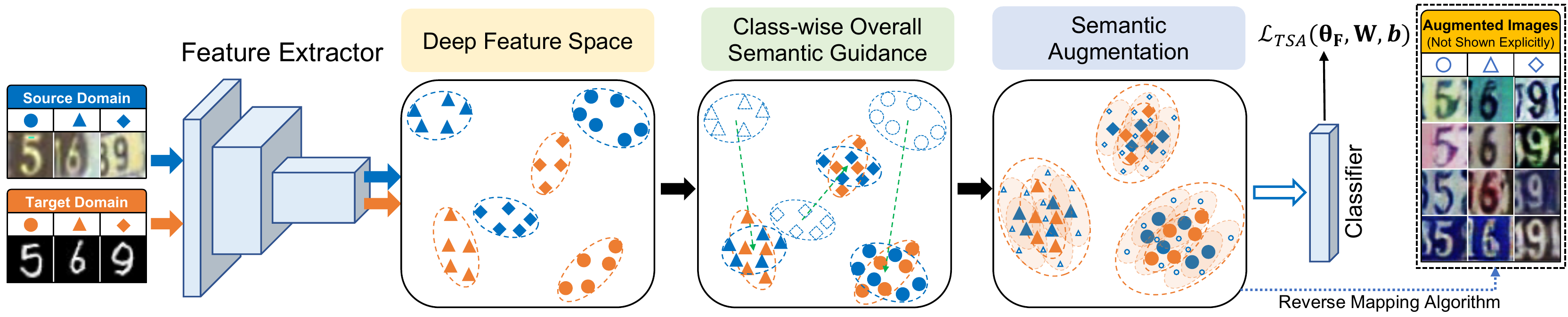}
    \caption{Illustration of TSA, where blue and orange dots represent source and target data respectively. For each class, we exploit the inter-domain feature mean difference which is denoted as the green dashed arrow and target intra-class covariance to augment source features towards target style. Instead of explicitly generating augmented features, we achieve the source augmentation by minimizing the loss $\mathcal{L}_{TSA}$, which adapts the classifier from source to target. Moreover, we provide the visualization of augmented features using the reverse mapping algorithm in the supplement. The augmented source data with different shapes, colors and backgrounds manifest that TSA can perform meaningful semantic transformations.}
    \label{Fig_TSDA}
    \vspace{-4mm}
\end{figure*}

\subsection{Classifier Adaptation}

On par with feature adaptation methods, classifier adaptation is also an indispensable part of DA, since the assumption that source and target domains can share one identical classifier is rather restrictive in practical scenarios.

A branch of classifier adaptation methods augments training samples towards target style to adapt the classifier \cite{SEDASS, DADA, CyCADA, AFAUDA, TAT, UNIT, GVB}. To be specific, Hoffman et al. \cite{CyCADA} exploit pixel cycle-consistency to make transformed images visually close to target domain images and use transformed samples to train the target model. UNIT \cite{UNIT} exploits image-to-image translation framework based on coupled GANs to reconstruct images. TAT \cite{TAT} generates augmented features with reversed gradients to adversarially train the classifier. \cite{GVB} leverages the gradually vanishing bridge (GVB) mechanism to generate intermediate features to bridge two domains.

Apart from the generative methods, some methods do not require to explicitly generate samples \cite{RTN, SWD, SymNets}. For instance, RTN \cite{RTN} plugs a residual block right after the source classifier to model and mitigate the classifier perturbation. SymNets \cite{SymNets} achieves joint distribution alignment by constructing three different task-specific classifiers.

By contrast, TSA does not need to delicately design auxiliary network modules or explicitly generate training samples. TSA implicitly generates augmented source features with target semantics (e.g., object shapes, colors or backgrounds) to adapt the classifier only by minimizing a robust transferable cross-entropy loss, which is extremely efficient.

\section{Transferable Semantic Augmentation}

\subsection{Motivation and Preliminaries}

This work delves into the classifier adaptation for DA problems, which has not yet been fully explored. We propose a Transferable Semantic Augmentation (TSA) approach to effectively augment source features towards target semantics (e.g., backgrounds, view angles or colors) by generating a very large (infinite) number of meaningful semantic transformation directions. The augmented source features will facilitate adapting the classifier from source to target successfully. TSA is inspired by the fact that deep neural networks surprisingly excel at linearizing features of input data, making it linearly separable \cite{DFI,ISDA}. Therefore, the relative positions in deep feature space can represent certain semantic transformations in the original input space. Based on this observation, MCF \cite{MCF} and ISDA \cite{ISDA} have performed semantic augmentation for linear models and deep networks, respectively. However, neither of them considers the crucial domain shift issue in the real world.

To tackle the cross-domain semantic mismatch problems, we need to perform meaningful and target-wise semantic augmentations for source data. In other words, the semantic augmentation directions should effectively guide source features to be target style, such as the shape, color, visual angle or background of target data. 

Formally, we denote the source and target domains in DA as $\mathcal{S}=\{(\x_{si}, y_{si})\}_{i=1}^{n_{s}}$ with $n_{s}$ labeled samples and $\mathcal{T}=\{\x_{tj}\}_{j=1}^{n_{t}}$ with $n_{t}$ unlabeled samples, respectively, where $y_{si}\in\{1,2, ... ,C\}$ is the label of $\x_{si}$. Note that, source and target data are sampled from different distributions, such that certain discrepancy may exist between their corresponding ideal classifiers. Our goal is to adapt the source-supervised classifier to target by augmenting the source feature representation $\mathbf{f}_{s}=F(\x_{s}, \boldsymbol{\Theta}_{F})\in\mathbb{R}^{K}$, where $K$ is the number of feature embedding dimensions and $F(\cdot)$ is the feature extractor parameterized by $\boldsymbol{\Theta}_{F}$. 

\subsection{Cross-Domain Semantic Augmentation}

As aforementioned, certain translating directions in deep feature space represent meaningful semantic transformations in the original input space. Thus, we attempt to augment source data towards target in deep feature space. However, manually searching for such cross-domain transformation directions is nontrivial. To address the issue, for each class, we propose to sample random vectors from a multivariate normal distribution with inter-domain feature mean difference as mean and class-conditional target covariance as the covariance matrix. By doing so, rich meaningful cross-domain transformation directions can be discovered. Then, we design an expected transferable cross-entropy loss over the augmented feature distribution to effectively and efficiently adapt the classifier from source to target domain.

\vspace{-2mm}
\subsubsection{Class-wise Overall Semantic Guidance.}

Due to the domain shift, for the same class, the means of source and target samples are different in the deep feature space. Such difference reflects the bias of the overall semantics, an integration of various semantics in one class. To conduct useful transformations, we force the mean alignment between augmented source features and target features class-wisely, which effectively bridges the overall semantic bias across two domains. Specifically, for each class $c$, 
let $\boldsymbol{\mu}_{s}^{c}$ and $\boldsymbol{\mu}_{t}^{c}$ denote the estimated mean feature vector for source and target domains, respectively. Here, pseudo labels for target data are employed as $y^\prime_{tj}=\mathop{\arg\max}_{c}{P^c_{tj}}$ to address the label scarce problem in target domain, where $\boldsymbol{P}_{tj}$ is the softmax outputs of target sample $\boldsymbol{x}_{tj}$. Then, the inter-domain mean difference $\boldsymbol{\Delta\mu}^{c} = \boldsymbol{\mu}_{t}^{c} -\boldsymbol{\mu}_{s}^{c}$ can be exploited to mitigate such overall semantic bias.

\vspace{-2mm}
\subsubsection{Semantic Transformation Direction Learning.}

To facilitate meaningful cross-domain semantic augmentations, all possible feature transformation directions towards target style need to be discovered. However, only utilizing the inter-domain mean difference is infeasible to fully explore target semantic variations.

Therefore, to address this problem, we further exploit the intra-class covariance of target features to capture target semantic variations. Specifically, we randomly sample semantic transformation directions from a multivariate distribution $N(\boldsymbol{\Delta\mu}^{c}, \boldsymbol{\Sigma}^{c}_{t})$ for each class $c$, and $\boldsymbol{\Sigma}^{c}_{t}$ is the corresponding target intra-class covariance, which contains rich target semantic variations. Here, we develop the normal distribution class-wisely, since the semantic knowledge in each class may differ from each other vastly. Notably, $\boldsymbol{\Delta\mu}^{c}$ attends to mitigating the overall semantic bias for class $c$, and $\boldsymbol{\Sigma}^{c}_{t}$ focuses on providing abundant target intra-class semantic variation knowledge. By adding the sampled transformation vectors, the augmented source features will be close to target domain and vary along target semantic variations.

\vspace{-2mm}
\subsubsection{Memory Module.}

To efficiently implement TSA in an end-to-end training manner, for each class $c$, we propose to estimate $\boldsymbol{\Delta\mu}^{c}$ and $\boldsymbol{\Sigma}^{c}_{t}$ according to a memory module that stores all the latest features. By contrast, ISDA \cite{ISDA} proposes an iterative manner by accumulating statistics of batches from first to current batch. However, the weight distribution of network in early stage will vastly differ from that in later stage. Thus, due to its accumulation property, such iterative manner may encounter the issue that out-of-date features will bias the estimation of mean and covariance. To avoid this, we use a memory module to cache all features of two domains and update in each batch. By doing so, we can discard those out-of-date features and replace them with the latest ones at the negligible cost of memory. Formally, in each iteration $i$, we will update a batch of  features and corresponding labels in memory module $\mathbb{M}$:

\vspace{-3mm}
\begin{small}
\begin{align}
    \mathbf{f}^{\mathbb{M}}_{j} \leftarrow \mathbf{f}^{(i)}_{j}, y^{\mathbb{M}}_{j} \leftarrow {y^\prime}^{(i)}_{j}, j \in \B^{(i)},
\end{align}\end{small}where $j$ is the sample index within a batch $\B^{(i)}$ and $\mathbf{f}^{\mathbb{M}}_{j}$/$y^{\mathbb{M}}_{j}$ is the feature/label stored in memory module $\mathbb{M}$. Comparisons of the two manners will be shown in the experiment.

\vspace{-1mm}
\subsubsection{Sampling Strategy.}

In the early training stage, the predictions of target samples are inevitably not that accurate. Thus, the estimation of target mean and covariance may not be quite informative as expected. To alleviate this issue, for each class $c$, we sample semantic transformation directions from $N(\lambda\boldsymbol{\Delta\mu}^{c}, \lambda\boldsymbol{\Sigma}^{c}_{t})$, where $\lambda$ is a positive parameter to control the augmentation strength. Since the target predictions will become more and more accurate as the training progresses, here we set $\lambda = (t/T)\times{\lambda_{0}}$, where $t$ and $T$ are the current and maximum iterations respectively, and $\lambda_{0}$ is a hyper-parameter. Consequently, as the training goes on, $\lambda$ will gradually grow from 0 to $\lambda_{0}$. And this sampling strategy will reduce the impact of less accurate estimations of mean and covariance at the beginning of the training stage.

\subsection{Transferable Cross-Entropy Loss Learning}

Once $C$ sampling distributions are constructed, each source deep feature $\mathbf{f}_{si}$ can conduct various semantic transformations along the random directions sampled from $N(\lambda\boldsymbol{\Delta\mu}^{y_{si}}, \lambda\boldsymbol{\Sigma}^{y_{si}}_{t})$ to generate the augmented feature $\tilde{\mathbf{f}}_{si}$, i.e., $\tilde{\mathbf{f}}_{si} \sim N(\mathbf{f}_{si}+\lambda \boldsymbol{\Delta\mu}^{y_{si}}, \lambda\boldsymbol{\Sigma}^{y_{si}}_{t})$. Considering a naive method, we can augment each $\mathbf{f}_{si}$ for $M$ times with its label preserved, which will result in an augmented feature set $\{(\mathbf{f}_{si}^{1}, y_{si}), (\mathbf{f}_{si}^{2}, y_{si}),...,(\mathbf{f}_{si}^{M}, y_{si})\}_{i=1}^{n_{s}}$. Based on this, the source network can be trained with a traditional cross-entropy loss on the augmented feature set:

\vspace{-3mm}
\begin{scriptsize}
\begin{align}
    {\mathcal{L}}_{M}(\boldsymbol{\Theta}_{F}, \mathbf{W}, \boldsymbol{b}) = \frac{1}{n_{s}}\sum_{i=1}^{n_{s}}\frac{1}{M}\sum_{m=1}^{M}-\log\left(\frac{e^{\boldsymbol{w}_{y_{si}}^{\top}\mathbf{f}_{si}^{m}+b_{y_{si}}}}{\sum_{c=1}^{C}e^{\boldsymbol{w}_{c}^{\top}\mathbf{f}_{si}^{m}+b_{c}}}\right),
\end{align}
\end{scriptsize}where $\mathbf{W} = [\boldsymbol{w}_{1}, \boldsymbol{w}_{2}, ..., \boldsymbol{w}_{C}]^{\top} \in \mathbb{R}^{C\times{K}}$ and $\boldsymbol{b} = [b_{1}, b_{2}, ..., b_{C}]^{\top}\in \mathbb{R}^{C}$ are the weight matrix and bias vector of the last fully connected layer, respectively. 

To achieve desired performance, $M$ is usually large, resulting in unexpected computational cost. In TSA, instead of explicitly generating the augmented source features for $M$ times, we intend to implicitly generate infinite augmented source features. When $M$ approaches infinity, we can derive an upper-bound loss according to the Law of Large Numbers over the augmented source distribution. 

Specifically, in the case of $M\rightarrow\infty$, our proposed expected transferable cross-entropy loss over the augmented source feature distribution is defined as:

\vspace{-2mm}
\begin{scriptsize}
\begin{align}
    \label{Eq:transferable_CE_loss}
    \lim_{M\rightarrow \infty}{\mathcal{L}}_{M} 
    &= \frac{1}{n_{s}}\sum_{i=1}^{n_{s}}\mathbb{E}_{\tilde{\mathbf{f}}_{si}}\left[-\log\left(\frac{e^{\boldsymbol{w}_{y_{si}}^{\top}\tilde{\mathbf{f}}_{si}+b_{y_{si}}}}{\sum_{c=1}^{C}e^{\boldsymbol{w}_{c}^{\top}\tilde{\mathbf{f}}_{si}+b_{c}}}\right)\right]\notag\\ 
    &= \frac{1}{n_{s}}\sum_{i=1}^{n_{s}}\mathbb{E}_{\tilde{\mathbf{f}}_{si}}\left[\log\left(\sum_{c=1}^{C}e^{(\boldsymbol{w}_{c}^{\top}-\boldsymbol{w}_{y_{si}}^{\top})\tilde{\mathbf{f}}_{si}+(b_{c}-b_{y_{si}})}\right)\right].
\end{align}
\end{scriptsize}
\vspace{-2mm}

However, it is infeasible to tackle Eq \eqref{Eq:transferable_CE_loss} directly. According to the Jensen's inequality \cite{statistical} and the concave property of logarithmic function $\log(\cdot)$, we have $\mathbb{E}[\log(X)] \leq \log(\mathbb{E}[X])$. Thus, we can derive the upper bound of the expected loss as follows:

\vspace{-4mm}
\begin{scriptsize}
\begin{align}
    \lim_{M\rightarrow \infty}{\mathcal{L}}_{M}
    &\leq \frac{1}{n_{s}}\sum_{i=1}^{n_{s}}\log\left(\mathbb{E}_{\tilde{\mathbf{f}}_{si}}\left[\sum_{c=1}^{C}e^{(\boldsymbol{w}_{c}^{\top}-\boldsymbol{w}_{y_{si}}^{\top})\tilde{\mathbf{f}}_{si}+(b_{c}-b_{y_{si}})}\right]\right)\notag\\
    &= \frac{1}{n_{s}}\sum_{i=1}^{n_{s}}\log\left(\sum_{c=1}^{C}\mathbb{E}_{\tilde{\mathbf{f}}_{si}}\left[e^{(\boldsymbol{w}_{c}^{\top}-\boldsymbol{w}_{y_{si}}^{\top})\tilde{\mathbf{f}}_{si}+(b_{c}-b_{y_{si}})}\right]\right).
\end{align}
\end{scriptsize}
\vspace{-2mm}

Due to $\tilde{\mathbf{f}}_{si}\sim N(\mathbf{f}_{si}+\lambda\boldsymbol{\Delta\mu}^{y_{si}}, \lambda\boldsymbol{\Sigma}^{y_{si}}_{t})$, we can derive that 
$(\boldsymbol{w}_{c}^{\top}-\boldsymbol{w}_{y_{si}}^{\top})\tilde{\mathbf{f}}_{si}+(b_{c}-b_{y_{si}})\sim N((\boldsymbol{w}_{c}^{\top}-\boldsymbol{w}_{y_{si}}^{\top})(\mathbf{f}_{si}+\lambda\boldsymbol{\Delta\mu}^{y_{si}})+(b_{c}-b_{y_{si}}), \sigma^{c}_{si})$, 
where $\sigma^{c}_{si}=\lambda(\boldsymbol{w}_{c}^{\top}-\boldsymbol{w}_{y_{si}}^{\top})\boldsymbol{\Sigma}^{y_{si}}_{t}(\boldsymbol{w}_{c}-\boldsymbol{w}_{y_{si}})$. Leveraging the moment-generating function $\mathbb{E}[e^{aX}] = e^{a\mu+\frac{1}{2}a^{2}\sigma}$,  $X\sim N(\mu, \sigma)$, we can obtain
\vspace{-2mm}
\begin{align}
    \label{Eq:loss_upper_bound}
    \lim_{M\rightarrow \infty}{\mathcal{L}}_{M} & \leq {\mathcal{L}}_{\infty} = - \frac{1}{n_{s}}\sum_{i=1}^{n_{s}}\log\frac{e^{{Z}^{y_{si}}_{si}}}{\sum_{c=1}^{C}e^{{Z}^{c}_{si}}},
\end{align}where ${Z}^{c}_{si} = \hat{y}_{si}^{c}+\lambda(\boldsymbol{w}_{c}^{\top}-\boldsymbol{w}_{y_{si}}^{\top})\boldsymbol{\Delta\mu}^{y_{si}}+\frac{\sigma^{c}_{si}}{2}$ and $\hat{y}_{si}^{c}$ is the $c$-th element of logits output of $\x_{si}$. 

Essentially, Eq \eqref{Eq:loss_upper_bound} gives us a surrogate loss ${\mathcal{L}}_{\infty}$ for optimizing the expected transferable cross-entropy loss, which could effectively adapt the source classifier to target with the augmented source features. Moreover, the proposed novel loss allows us to utilize TSA as a plug-in module for other DA methods to further improve their transferability.

\textbf{Discussion.} Although the derivation of the upper bound is kind of similar with ISDA \cite{ISDA}, the motivation of our method is essentially different. We aim to enhance the classifier adaptation for DA where target domain is unlabeled and large domain shift exists between domains. Thus, the proposed $\boldsymbol{\Delta\mu}^{c}$ and $\boldsymbol{\Sigma}^{c}_{t}$ are crucial to successfully generate transformation directions that can really bridge the domain gap. While ISDA is a completely supervised algorithm and ignores the crucial domain shift issues in reality. In the ablation study, we show that the performance of directly applying ISDA in DA context is inferior. Besides, different from the iterative estimation manner in ISDA, our memory module manner can achieve more accurate estimations of the mean and covariance, which is shown in the experiment.

\subsection{Overall Formulation}

In information theory, mutual information $I(X;Y)$ measures how related two random variables $X$ and $Y$ are. Actually, strong correlations between target features and predictions will benefit our semantic augmentations, because the extracted features will be more informative and contain more important semantics for predictions, ignoring trivial semantics. Thus, we employ the mutual information maximization on target data, i.e., minimizing the loss in Eq \eqref{eq:MI}.

\begin{footnotesize}
\begin{align}
\label{eq:MI}
\mathcal{L}_{MI} = \sum_{c=1}^{C}\hat{P}^{c} \log \hat{P}^{c} - \frac{1}{n_{t}}\sum_{j=1}^{n_{t}}\sum_{c=1}^{C}{P_{tj}^{c}} \log P_{tj}^{c},
\end{align}
\end{footnotesize}
where $\boldsymbol{\hat{P}} = \frac{1}{n_{t}} \sum_{j=1}^{n_t} \boldsymbol{P}_{tj}$. Since target domain is unlabeled, we use the average of target predictions to approximate the ground-truth distribution on target domain.

To sum up, the overall objective function of TSA is:
\begin{align}
    \label{Eq:loss_all}
    \mathcal{L}_{TSA}&={\mathcal{L}}_{\infty}+\beta\mathcal{L}_{MI},
\end{align}where $\beta$ is a trade-off parameter. The effects of different parts in TSA will be analyzed in details in ablation study.

\subsection{Theoretical Insight}

To theoretically understand TSA, we introduce the domain adaptation theory proposed by \cite{A-distance}, which reveals the ingredients of target generalization error $\epsilon_{t}$. Formally, let $\mathcal{H}$ denote the hypothesis space and $h \in \mathcal{H}$ denote the classifier, we can formulate the upper bound of $\epsilon_{t}$ as:
\begin{align}
     \epsilon_{t}(h) \leq \epsilon_{s}(h) + \frac{1}{2} d_{\mathcal{H}\Delta\mathcal{H}}(\mathcal{S}, \mathcal{T}) + \lambda^{*},
     \forall h \in \mathcal{H},
\end{align}where $\epsilon_{s}(h)$ is the source generalization error of $h$, $d_{\mathcal{H}\Delta\mathcal{H}}(\mathcal{S}, \mathcal{T})$ is the $\mathcal{H}\Delta\mathcal{H}$-distance between source and target domains, and $\lambda^{*} = \epsilon_{s}(h^{*}) + \epsilon_{t}(h^{*})$ denotes the error of an ideal joint hypothesis $h^{*}$ on source and target domains. 

With the supervision of labeled source data, $\epsilon_{s}(h)$ is well bounded. Besides, TSA generates transformation directions from the constructed multivariate distribution to augment source features towards target domain. Since these augmented source features can fill the domain gap, TSA further bounds the $d_{\mathcal{H}\Delta\mathcal{H}}(\mathcal{S}, \mathcal{T})$. 
Moreover, TSA implicitly generates infinite augmented source features  that are close to target domain class-wisely, enabling the classifier to jointly minimize $\epsilon_{s}(h^{*})$ and $\epsilon_{t}(h^{*})$ of the shared error $\lambda^{*}$ on the augmented training set. To sum up, TSA complies well with the theory, thus further enhancing the transferability.

\begin{table*}[htbp]
  \small
  \setlength{\abovecaptionskip}{0.cm}
  \setlength{\belowcaptionskip}{0.cm}
  \centering
  \caption{Accuracy (\%) on Office-Home for UDA (ResNet-50).}
  \setlength{\tabcolsep}{0.4mm}{
    \begin{tabular}{|l|cccccccccccc|c|}
    \hline
    Method & Ar$\rightarrow$Cl & Ar$\rightarrow$Pr & Ar$\rightarrow$Rw & Cl$\rightarrow$Ar & Cl$\rightarrow$Pr & Cl$\rightarrow$Rw & Pr$\rightarrow$Ar & Pr$\rightarrow$Cl & Pr$\rightarrow$Rw & Rw$\rightarrow$Ar & Rw$\rightarrow$Cl & Rw$\rightarrow$Pr & Avg \\
    \hline
    \hline
    JAN \cite{JAN} & 45.9 & 61.2 & 68.9 & 50.4 & 59.7 & 61.0 & 45.8 & 43.4 & 70.3 & 63.9 & 52.4 & 76.8 & 58.3 \\
    TAT \cite{TAT} & 51.6 & 69.5 & 75.4 & 59.4 & 69.5 & 68.6 & 59.5 & 50.5 & 76.8 & 70.9 & 56.6 & 81.6 & 65.8 \\
    TPN \cite{TPN} & 51.2 & 71.2 & 76.0 & 65.1 & 72.9 & 72.8 & 55.4 & 48.9 & 76.5 & 70.9 & 53.4 & 80.4 & 66.2 \\
    ETD \cite{ETD} & 51.3 & 71.9 & \textbf{85.7} & 57.6 & 69.2 & 73.7 & 57.8 & 51.2 & 79.3 & 70.2 & 57.5 & 82.1 & 67.3 \\
    SymNets \cite{SymNets} & 47.7 & 72.9 & 78.5 & 64.2 & 71.3 & 74.2 & 64.2 & 48.8 & 79.5 & 74.5 & 52.6 & 82.7 & 67.6 \\
    BNM \cite{BNM} & 52.3 & 73.9 & 80.0 & 63.3 & 72.9 & 74.9 & 61.7 & 49.5 & 79.7 & 70.5 & 53.6 & 82.2 & 67.9 \\
    MDD \cite{MDD} & 54.9 & 73.7 & 77.8 & 60.0 & 71.4 & 71.8 & 61.2 & 53.6 & 78.1 & 72.5 & 60.2 & 82.3 & 68.1 \\
    GSP \cite{GSP} & 56.8 & 75.5 & 78.9 & 61.3 & 69.4 & 74.9 & 61.3 & 52.6 & 79.9 & 73.3 & 54.2 & 83.2 & 68.4 \\
    GVB-GD \cite{GVB} & 57.0 & 74.7 & 79.8 & \textbf{64.6} & 74.1 & 74.6 & 65.2 & 55.1 & 81.0 & 74.6 & 59.7 & 84.3 & 70.4 \\
    \hline
    ResNet-50 \cite{resnet}& 34.9 & 50.0 & 58.0 & 37.4 & 41.9 & 46.2 & 38.5 & 31.2 & 60.4 & 53.9 & 41.2 & 59.9 & 46.1 \\
    +TSA & 53.6 & 75.1 & 78.3 & 64.4 & 73.7 & 72.5 & 62.3 & 49.4 & 77.5 & 72.2 & 58.8 & 82.1 & 68.3 \\
    \hline
    DANN \cite{DANN} & 45.6 & 59.3 & 70.1 & 47.0 & 58.5 & 60.9 & 46.1 & 43.7 & 68.5 & 63.2 & 51.8 & 76.8 & 57.6 \\
    +TSA & 56.5 & 71.5 & 78.7 & 62.2 & 73.5 & 72.9 & 61.8 & 55.6 & 80.6 & 72.6 & 61.3 & 82.0 & 69.1 \\
    \hline
    CDAN \cite{CDAN} & 50.7 & 70.6 & 76.0 & 57.6 & 70.0 & 70.0 & 57.4 & 50.9 & 77.3 & 70.9 & 56.7 & 81.6 & 65.8 \\
    +TSA & 56.7 & 75.3 & 80.5 & 63.9 & 75.6 & \textbf{75.5} & 63.8 & 55.5 & \textbf{81.6} & 74.8 & 60.2 & \textbf{84.4} & 70.7 \\
    \hline
    BSP \cite{BSP} & 52.0 & 68.6 & 76.1 & 58.0 & 70.3 & 70.2 & 58.6 & 50.2 & 77.6 & 72.2 & 59.3 & 81.9 & 66.3 \\
    +TSA & \textbf{57.6} & \textbf{75.8} & 80.7 & 64.3 & \textbf{76.3} & 75.1 & \textbf{66.7} & \textbf{55.7} & 81.2 & \textbf{75.7} & \textbf{61.9} & 83.8 & \textbf{71.2} \\
    \hline
    \end{tabular}}
  \label{tab:Office-Home}
  \vspace{-2mm}
\end{table*}

\section{Experiment}

\subsection{Datasets}

\textbf{Office-Home} \cite{Office-Home} is a challenging benchmark for domain adaptation, which contains 15,500 images in 65 classes drawn from 4 distinct domains: Artistic (Ar), Clip Art (Cl), Product (Pr), and Real-World (Rw).

\textbf{Office-31} \cite{office-31} is a classical cross-domain benchmark, including images in 31 classes drawn from 3 distinct domains: Amazon (A), Webcam (W) and DSLR (D).

\textbf{VisDA-2017} \cite{visda2017} is a large-scale dataset for visual domain adaptation, containing over 280K images across 12 categories. Following \cite{MCD}, we choose the synthetic domain with 152,397 images as source and the realistic domain with 55,388 images as target.

\textbf{Digital Datasets} contain 4 standard digital datasets: MNIST \cite{MNIST}, USPS \cite{USPS}, Street View House Numbers (SVHN) \cite{svhn} and synthetic digits dataset (SYN) \cite{syn}. All of these datasets provide number images from 0 to 9. We construct four transfer tasks: MNIST to USPS (M $\rightarrow$ U), USPS to MNIST (U $\rightarrow$ M), SVHN to MNIST (SV $\rightarrow$ M) and SYN to MNIST (SY $\rightarrow$ M).

\subsection{Implementation Details}
For a fair comparison, we use ResNet \cite{resnet} pre-trained on ImageNet \cite{imagenet2014} as the backbone network for datasets: Office-Home, Office-31 and VisDA-2017. As for digital datasets, we employ the same network structures in \cite{JADA} and train the networks from scratch. In this paper, all experiments are implemented via PyTorch \cite{paszke2019pytorch}. We adopt mini-batch SGD optimizer with momentum $0.9$ for network optimization, and deep embedded validation \cite{dev} to select hyper-parameters $\lambda_0$ from $\{0.1, 0.25, 0.5, 0.75, 1.0\}$ and $\beta$ from $\{0.01, 0.05, 0.1, 0.15, 0.2\}$, and we found $\lambda_0=0.25, \beta=0.1$ works well on all datasets. 

We evaluate our approach by applying TSA to the source-only model and several mainstream DA methods, i.e., DANN \cite{DANN}, CDAN \cite{CDAN}, BSP \cite{BSP} and JADA \cite{JADA}, based on their open-source codes. For each task, we report the average accuracy of 3 random trials. Code of TSA is available at \url{https://github.com/BIT-DA/TSA}.

\begin{table}[htbp]
  \small
  \setlength{\abovecaptionskip}{0.cm}
  \setlength{\belowcaptionskip}{0.cm}
  \centering
  \caption{Accuracy (\%) on Office-31 for UDA (ResNet-50).}
    \setlength{\tabcolsep}{0.5mm}{
    \begin{tabular}{|l|cccccc|c|}
    \hline
    Method & A$\rightarrow$W & D$\rightarrow$W & W$\rightarrow$D & A$\rightarrow$D & D$\rightarrow$A & W$\rightarrow$A & Avg \\
    \hline
    \hline
    ADDA \cite{ADDA} & 86.2 & 96.2 & 98.4 & 77.8 & 69.5 & 68.9 & 82.9 \\
    JAN \cite{JAN} & 85.4 & 97.4 & 99.8 & 84.7 & 68.6 & 70.0 & 84.3 \\
    ETD \cite{ETD} & 92.1 & \textbf{100.0} & \textbf{100.0} & 88.0 & 71.0 & 67.8 & 86.2 \\
    MCD \cite{MCD} & 88.6 & 98.5 & \textbf{100.0} & 92.2 & 69.5 & 69.7 & 86.5 \\
    BNM \cite{BNM} & 91.5 & 98.5 & \textbf{100.0} & 90.3 & 70.9 & 71.6 & 87.1 \\
    DMRL \cite{DMRL} & 90.8 & 99.0 & \textbf{100.0} & 93.4 & 73.0 & 71.2 & 87.9 \\
    SymNets \cite{SymNets} & 90.8 & 98.8 & \textbf{100.0} & 93.9 & 74.6 & 72.5 & 88.4 \\
    TAT \cite{TAT} & 92.5 & 99.3 & \textbf{100.0} & 93.2 & 73.1 & 72.1 & 88.4 \\
    MDD \cite{MDD} & 94.5 & 98.4 & \textbf{100.0} & 93.5 & 74.6 & 72.2 & 88.9 \\
    GVB-GD \cite{GVB} & 94.8 & 98.7 & \textbf{100.0} & 95.0 & 73.4 & 73.7 & 89.3 \\
    GSP \cite{GSP} & 92.9 & 98.7 & 99.8 & 94.5 & 75.9& 74.9 & 89.5 \\
    \hline
    ResNet-50 \cite{resnet} & 68.4 & 96.7 & 99.3 & 68.9 & 62.5& 60.7 & 76.1 \\
    +TSA & 94.8 & 99.1 & \textbf{100.0} & 92.6 & 74.9 & 74.4 & 89.3 \\
    \hline
    DANN \cite{DANN} & 82.0 & 96.9 & 99.1 & 79.7 & 68.2& 67.4 & 82.2 \\
    +TSA & 94.9 & 98.5 & \textbf{100.0} & 92.0 & 76.3 & 74.6 & 89.4 \\
    \hline
    CDAN \cite{CDAN} & 94.1 & 98.6 & \textbf{100.0} & 92.9 & 71.0 & 69.3 & 87.7  \\
    +TSA & 95.1 & 98.7 & \textbf{100.0} & 95.0 & 76.3 & 75.9 & 90.2 \\
    \hline
    BSP \cite{BSP} & 93.3 & 98.2 & \textbf{100.0} & 93.0 & 73.6 & 72.6 & 88.5 \\
    +TSA & \textbf{96.0} & 98.7 & \textbf{100.0} & \textbf{95.4} & \textbf{76.7} & \textbf{76.8} & \textbf{90.6} \\
    \hline
    \end{tabular}}
  \label{tab:Office-31}
  \vspace{-3mm}
\end{table}

\subsection{Results} 

\textbf{Results on Office-Home} are reported in Table \ref{tab:Office-Home}. Office-Home is a challenging dataset with large domain discrepancy for DA. TSA consistently improves the generalization ability of the four based methods. Specifically, TSA empowers BSP with 4.9\% improvement, achieving the highest accuracy 71.2\% on average. Based on these promising results, we can infer that TSA can stably enhance the transferability of classifiers on this difficult cross-domain dataset.

\textbf{Results on Office-31} are presented in Table \ref{tab:Office-31}. Compared with feature adaptation methods (e.g., JAN and DANN), ResNet-50+TSA surpasses them by a large margin, indicating that classifier adaptation is also indispensable for DA. In particular, TSA brings a large improvement of 7.2\% to DANN, revealing that TSA is complementary to previous DA methods. Besides, ResNet-50+TSA also surpasses recent classifier adaptation methods, e.g., SymNets and TAT, revealing that TSA can explore truly useful semantic information to better adapt the classifier.

\begin{figure*}[htbp]
  \vspace{-3mm}
  \setlength{\abovecaptionskip}{0.cm}
  \setlength{\belowcaptionskip}{0.cm}
  \centering
  \includegraphics[width=0.9\textwidth]{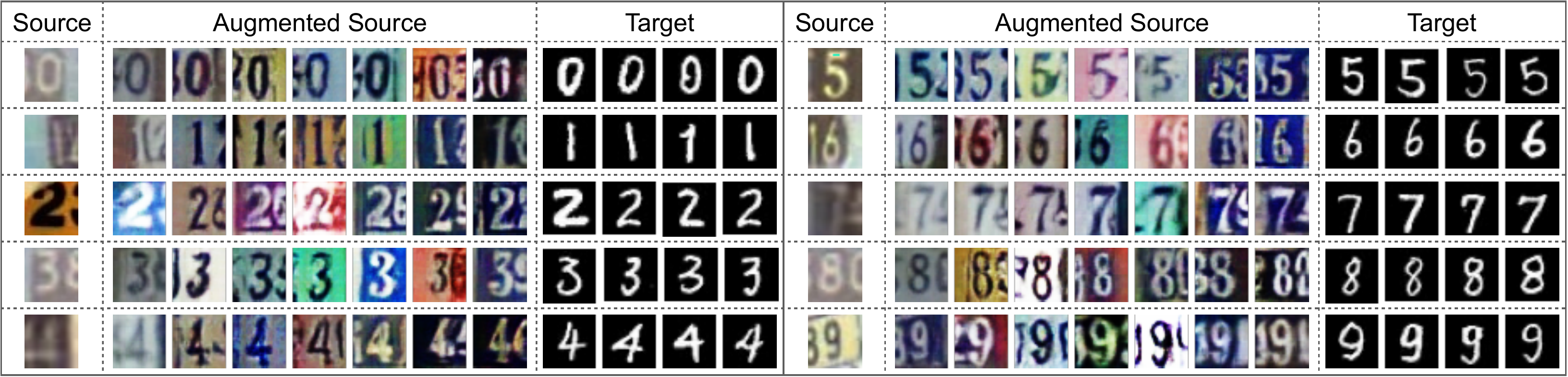}
  \caption{Visualization of semantically augmented images for task SVHN to MNIST. ``Source'' and ``Augmented Source'' columns present the images from SVHN and their corresponding augmented images visualized by using our designed reverse mapping algorithm (presented in the supplement), respectively. ``Target'' column provides several images representing corresponding classes from MNIST.}
  \label{Fig_visual_samples}
  \vspace{-3mm}
\end{figure*}

\textbf{Results on VisDA-2017} are summarized in Table \ref{tab:visda}. We can observe that ResNet-101+TSA dramatically outperforms other augmentation methods, like DMRL \cite{DMRL} and DM-ADA \cite{DM-ADA}. This is mainly due to the fact that TSA exploits the mean difference and target covariance to fully capture the meaningful semantic information class-wisely, achieving better augmentation results. Besides, TSA also brings remarkable improvements to other baseline methods, which validates the effectiveness and versatility of TSA.

\textbf{Results on Digital Datasets} are reported in Table \ref{tab:digital}. It is noteworthy that DIFA can be seen as DANN combined with extra GAN-based modules which are designed for generating training samples. Thus, it is fair to compare DIFA with DANN+TSA. DANN+TSA exceeds DIFA on average and especially surpasses it by a large margin on U ${\rightarrow}$ M. Besides, TSA does not require to introduce any auxiliary networks, which is extremely easy to implement. For CDAN and JADA, TSA also boosts their performance on average. These encouraging results validate that our semantic augmentation approach can derive a robust classifier efficiently.

\begin{table}[htbp]
  \small
  \setlength{\abovecaptionskip}{0.cm}
  \setlength{\belowcaptionskip}{0.cm}
  \centering
  \caption{Accuracy (\%) on VisDA-2017 for UDA (ResNet-101).}
    \begin{tabular}{|l|c|}
    \hline
    Method & Synthetic $\rightarrow$ Real  \\
    \hline
    \hline
    DAN \cite{DAN} & 61.1 \\
    MCD \cite{MCD} & 71.9 \\
    SimNet \cite{simnet}& 72.9 \\
    DMRL \cite{DMRL} & 75.5 \\
    DM-ADA \cite{DM-ADA} & 75.6 \\
    TPN \cite{TPN} & 80.4 \\
    \hline
    ResNet-101 \cite{resnet} & 52.4 \\
    +TSA & 78.6 (26.2 $\uparrow$)\\
    \hline
    DANN \cite{DANN} & 57.4 \\
    +TSA & 79.6 (22.2 $\uparrow$) \\
    \hline
    CDAN \cite{CDAN} & 73.7 \\
    +TSA & 81.6 (7.9 $\uparrow$)\\
    \hline
    BSP \cite{BSP} & 76.9 \\
    +TSA & \textbf{82.0} (5.1 $\uparrow$)\\
    \hline
    \end{tabular}
  \label{tab:visda}
  \vspace{-4mm}
\end{table}

\begin{table}[htbp]\scriptsize
  \small
  \setlength{\abovecaptionskip}{0.cm}
  \setlength{\belowcaptionskip}{0.cm}
  \centering
  \caption{Accuracy (\%) on Digital Datasets for UDA.}
  \setlength{\tabcolsep}{0.6mm}{
    \begin{tabular}{|l|cccc|c|}
    \hline
    Method & M $\rightarrow$ U &  U $\rightarrow$ M &  SV $\rightarrow$ M &  SY $\rightarrow$ M & Avg \\
    \hline
    \hline
    ADDA \cite{ADDA} & 89.4$\pm$0.2 & 90.1$\pm$0.8 & 76.0$\pm$1.8 & 96.3$\pm$0.4 & 88.0 \\
    PixelDA \cite{DADA} & 95.9$\pm$0.7 & - & - & - & - \\
    DIFA \cite{AFAUDA} & 92.3$\pm$0.1 & 89.7$\pm$0.5 & 89.7$\pm$2.0 & - & - \\
    UNIT \cite{UNIT} & 95.9$\pm$0.3 & 93.6$\pm$0.2 & 90.5$\pm$0.3 & - & - \\
    CyCADA \cite{CyCADA} & 95.6$\pm$0.2 & 96.5$\pm$0.1 & 90.4$\pm$0.4 & - & - \\
    TPN \cite{TPN} & 92.1$\pm$0.2 & 94.1$\pm$0.1 & 93.0$\pm$0.3 & - & \\
    DM-ADA \cite{DM-ADA} & 96.7$\pm$0.5 & 94.2$\pm$0.9 & 95.5$\pm$1.1 & - & - \\
    MCD \cite{MCD} & 96.5$\pm$0.3 & 94.1$\pm$0.3 & 96.2$\pm$0.4 & - & - \\
    ETD \cite{ETD} & 96.4$\pm$0.3 & 96.3$\pm$0.1 & 97.9$\pm$0.4 & - & \\
    DMRL \cite{DMRL} & 96.1$\pm$0.2 & \textbf{99.0$\pm$0.1} & 96.2$\pm$0.4 & - & - \\
    \hline
    Source-only & 79.4$\pm$0.4 & 63.4$\pm$0.3 & 67.1$\pm$0.5  & 89.7$\pm$0.2 & 74.9 \\
    +TSA & 90.9$\pm$0.3 & 96.9$\pm$0.2 & \textbf{99.2$\pm$0.1} & 97.8$\pm$0.2 & 96.2 \\
    \hline
    DANN \cite{DANN} & 85.1$\pm$0.5 & 73.0$\pm$0.2 & 71.1$\pm$0.4 & 90.2$\pm$0.2 & 79.9 \\
    +TSA & 93.2$\pm$0.3 & 97.7$\pm$0.2 & 97.1$\pm$0.3 & 98.0$\pm$0.2 & 96.5 \\
    \hline
    CDAN \cite{CDAN} & 95.6$\pm$0.1 & 98.0$\pm$0.1 & 89.2$\pm$0.3 & - & - \\
    +TSA & 95.1$\pm$0.2 & 98.7$\pm$0.1 & 98.0$\pm$0.1 & 98.2$\pm$0.2 & 97.5 \\
    \hline
    JADA \cite{JADA} & 97.6$\pm$0.2 & 97.1$\pm$0.3 & 96.4$\pm$0.2 & 98.6$\pm$0.2 & 97.4  \\
    +TSA & \textbf{98.0$\pm$0.1} & 98.3$\pm$0.3 & 98.7$\pm$0.2 & \textbf{99.2$\pm$0.1} & \textbf{98.5} \\
    \hline
    \end{tabular}}
  \label{tab:digital}
  \vspace{-1mm}
\end{table}

\begin{figure*}[htbp]\centering
  \vspace{-3mm}
  \setlength{\abovecaptionskip}{0.cm}
  \setlength{\belowcaptionskip}{0.cm}
  \centering
  \includegraphics[width=0.83\textwidth]{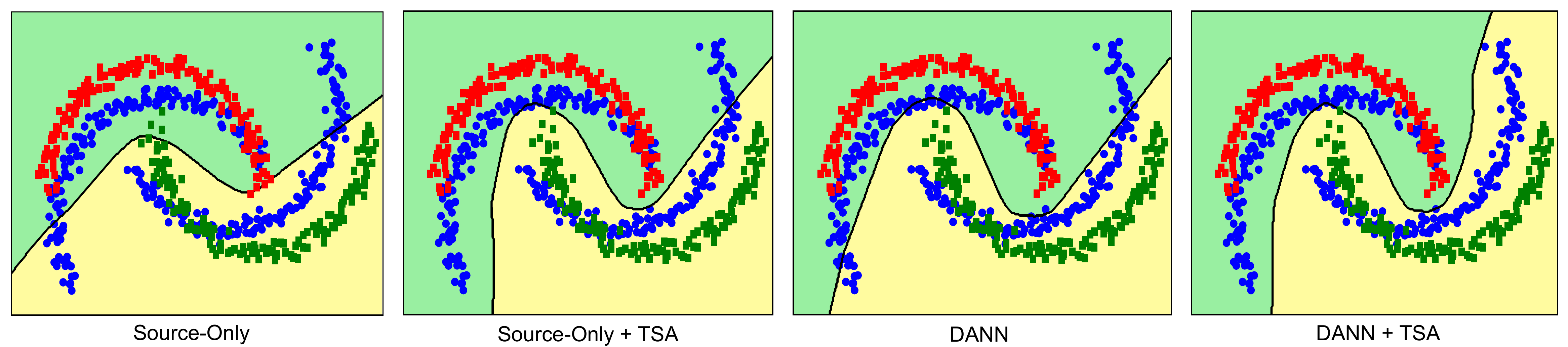}
  \caption{Red and green dots denote source data of class $0$ and $1$. Blue dots are target data, generated via rotating source data by $30^{\circ}$. Dots in green and yellow areas are respectively classified as $0$ and $1$. Black solid lines are classification boundaries.}
  \label{Fig_moon}
  \vspace{-3mm}
\end{figure*}

\begin{table}[htbp]
  \small
  \setlength{\abovecaptionskip}{0.cm}
  \setlength{\belowcaptionskip}{0.cm}
  \centering
  \caption{Ablation Study of TSA on Office-31.}
  \setlength{\tabcolsep}{0.28mm}{
    \begin{tabular}{|l|cccccc|c|}
    \hline
    method & A$\rightarrow$W  & D$\rightarrow$W  & W$\rightarrow$D  & A$\rightarrow$D  & D$\rightarrow$A  & W$\rightarrow$A  & Avg \\
    \hline
    \hline
    DANN \cite{DANN} & 82.0 & 96.9 & 99.1 & 79.7 & 68.2 & 67.4 & 82.2  \\
    +TSA (w/o $\boldsymbol{\Delta\mu}^{c}$) & 92.5 & 98.5 & 100.0 & 90.8 & 70.4 & 69.4 & 86.9 \\
    +TSA (w/o $\boldsymbol{\Sigma}_{t}^{c}$) & 92.3 & 98.8 & 100.0 & 91.1 & 71.8 & 70.9 & 87.5 \\
    \hline
    +$\mathcal{L}_{MI}$ & 91.6 & 98.2 & 100.0 & 89.8 & 69.7 & 68.9 & 86.4 \\
    +TSA (w/o $\mathcal{L}_{MI}$) & 92.7 & 98.5 & 100.0 & 90.6 & 70.7 & 70.6 & 87.2 \\
    +ISDA & 90.3 & 97.9 & 100.0 & 87.1 & 68.5 & 68.8 & 85.4 \\
    +TSA & 94.9 & 98.5 & 100.0 & 92.0 & 76.3 & 74.6 & \textbf{89.4} \\
    \hline
    \end{tabular}}
    \vspace{-5mm}
  \label{tab:ablation}
\end{table}

\subsection{Analysis}
\label{Sec:analysis}
\textbf{Ablation Study}. For each class $c$, the multivariate normal distribution for sampling transformation directions consists of two important parts: 1) inter-domain feature mean difference $\boldsymbol{\Delta\mu}^{c}$ and 2) target intra-class covariance matrix $\boldsymbol{\Sigma}_{t}^{c}$. Only using $\boldsymbol{\Sigma}_{t}^{c}$ to build sampling distribution $N(0, \lambda\boldsymbol{\Sigma}_{t}^{c})$ will suffer from overall semantic bias due to the natural dissimilarity between two domains. While only using $\boldsymbol{\Delta\mu}^{c}$ to construct the distribution $N(\lambda\boldsymbol{\Delta\mu}^{c}, 0)$ cannot fully capture semantic variations underlying target domain.

Therefore, to investigate the effects of different components in the sampling distribution, we conduct extensive ablation studies of TSA based on DANN \cite{DANN} in Table \ref{tab:ablation}. The results manifest that our strategy achieves the best performance, indicating that our sampling distribution can yield meaningful transformation directions to guide the augmented source features towards target.

Fixing the sampling distribution $N(\lambda\boldsymbol{\Delta\mu}^{c},\lambda\boldsymbol{\Sigma}_{t}^{c})$, we further explore the impacts of mutual information maximization loss $\mathcal{L}_{MI}$ in TSA. In Table \ref{tab:ablation}, TSA seems to be more effective when incorporating $\mathcal{L}_{MI}$, since mutual information maximization helps achieve more accurate target predictions and more informative target features, which is beneficial for capturing the underlying target semantics. Besides, DANN+TSA (w/o $\mathcal{L}_{MI}$) achieves much better results than DANN+ISDA. The reason is that ISDA \cite{ISDA} does not consider the domain shift problem and fails to capture target semantics, resulting in poor classifier adaptability.
 
\textbf{Visualization for Source Augmentation}. 
To intuitively validate that TSA can generate meaningful augmented features, we design a reverse mapping algorithm (presented in the supplement) to search images corresponding to the augmented features. From Fig. \ref{Fig_visual_samples}, we can observe that TSA can alter the semantics of source images, such as backgrounds, colors and shapes of numbers. Moreover, some augmented images closely resemble target images. This verifies that TSA can indeed generate diverse, meaningful and even target-specific augmented features. More augmented images are shown in the supplement.

\textbf{Inter Twinning Moons 2D Problems}. 
The goal of TSA is to learn transferable classifiers. We carry out experiments on inter twinning moons 2D problem \cite{scikit-learn} to illustrate the adaptation of classification boundaries brought by TSA. In Fig. \ref{Fig_moon}, source-only model and DANN correctly classify almost all source samples, but misclassify many target samples. By contrast, the classification boundaries of source-only+TSA and DANN+TSA are adapted to target data clearly. The apparent classifier movement reveals that TSA can enhance the classifier transferability notably.

\begin{figure}[htbp]
  \vspace{-3mm}
  \centering
  \subfigure[$\boldsymbol{\Delta\mu}^c$]{
  \begin{minipage}[t]{0.45\linewidth}
  \centering
  \includegraphics[width=1\linewidth]{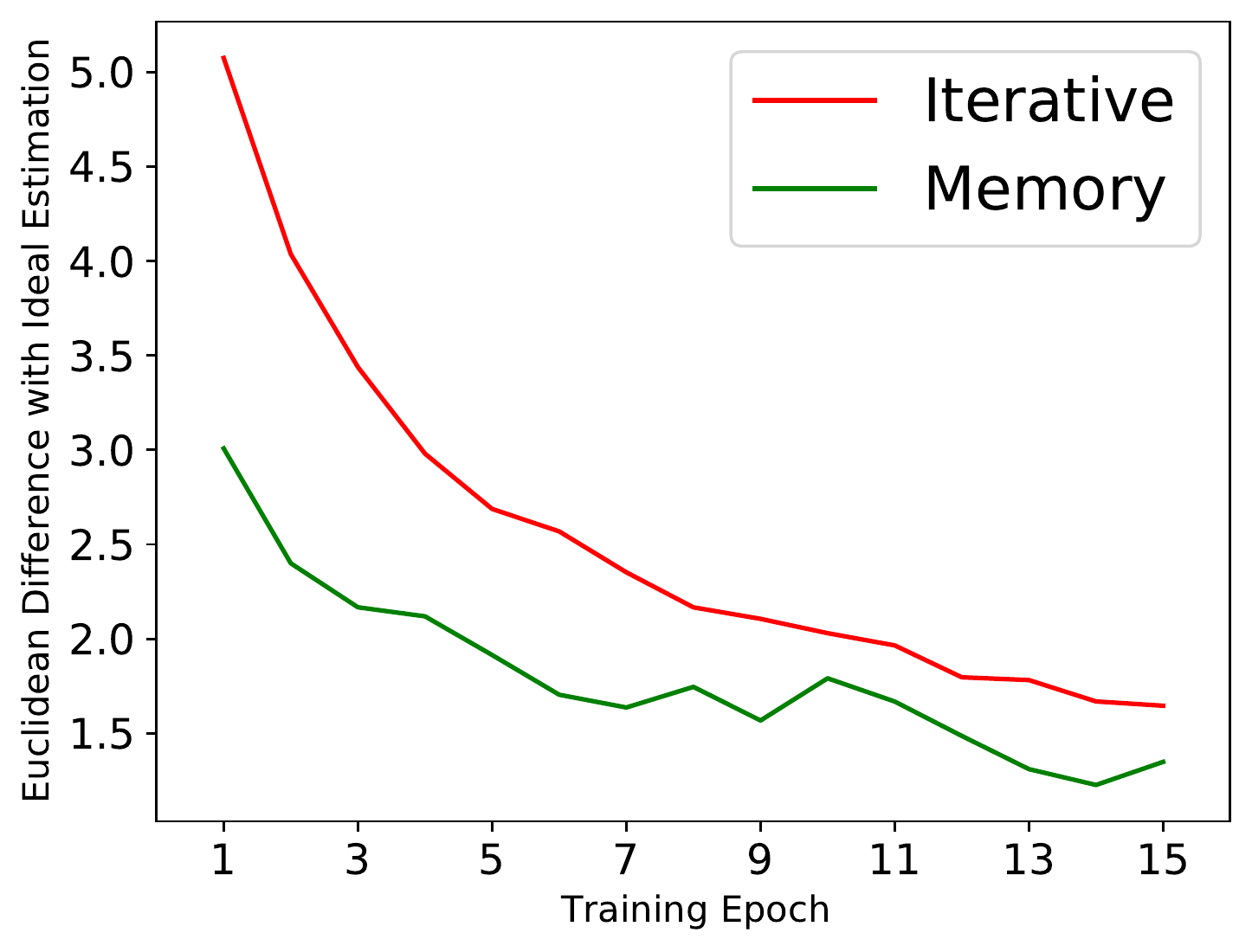}
  \end{minipage}
  }
  \subfigure[$\boldsymbol{\Sigma}_t^c$]{
  \begin{minipage}[t]{0.45\linewidth}
  \centering
  \includegraphics[width=1\linewidth]{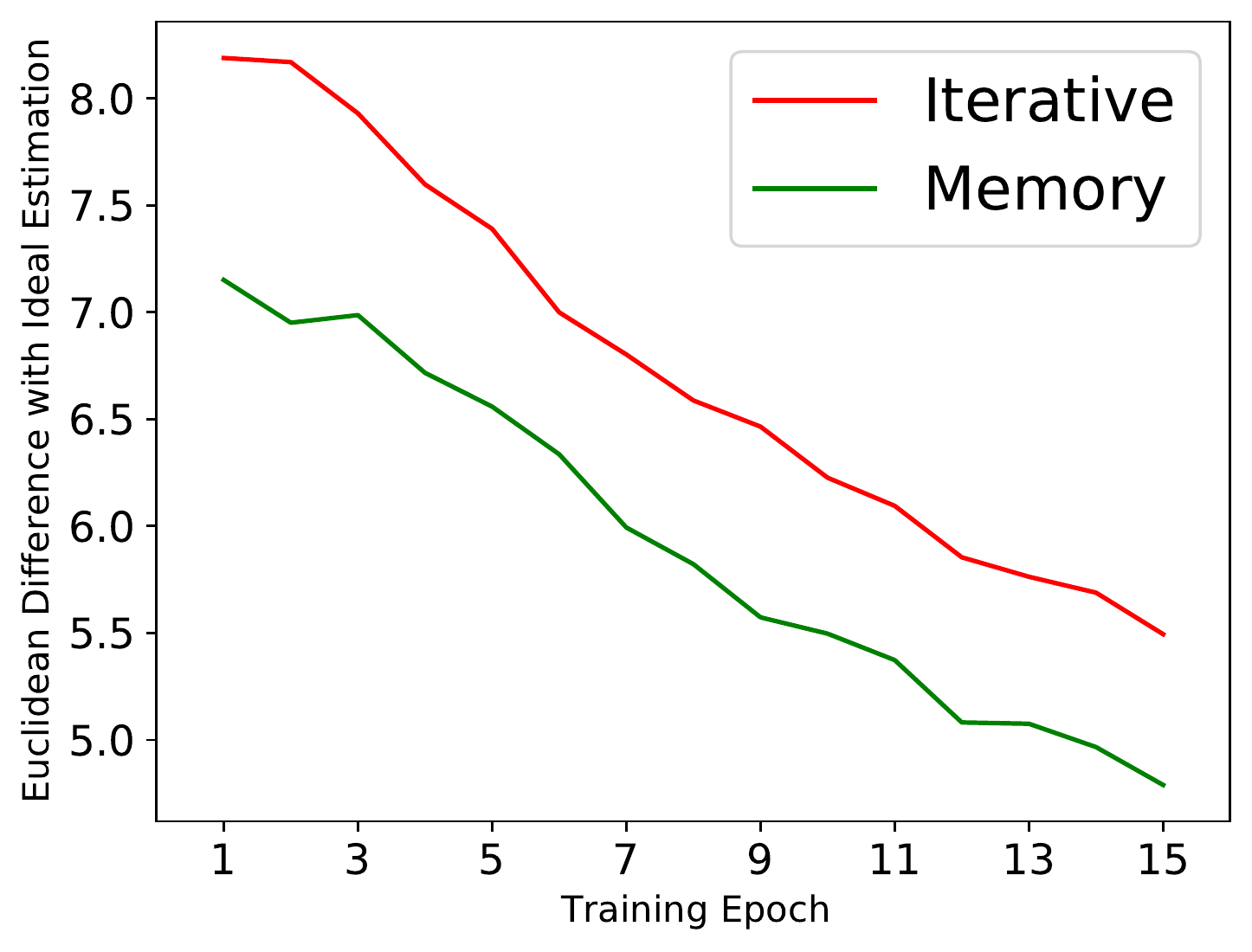}
\end{minipage}
}
\centering
\caption{Estimation bias of the iterative manner in \cite{ISDA} and our memory module manner on D$\rightarrow$A in different epochs. (a) and (b) present the average estimations bias of inter-domain mean difference $\boldsymbol{\Delta\mu}^c$ and target intra-class covariance $\boldsymbol{\Sigma}_t^c$ for all $C$ classes.}
\label{Fig_iter_mem}
\vspace{-3mm}
\end{figure}

\textbf{Estimation Bias Comparisons.} In this section, we compare the estimation bias of the iterative manner in ISDA \cite{ISDA} and our memory module manner. Specifically, in each epoch, we calculate the average Euclidean distances between ideal and practical estimations of the mean and covariance for all $C$ classes, respectively. The ideal estimation denotes using all features that are freshly produced by feeding the whole dataset into current network to estimate $\boldsymbol{\Delta\mu}^c$ and $\boldsymbol{\Sigma}_t^c$ for each class $c$; the practical estimation denotes leveraging the iterative manner in ISDA or our memory module manner to class-wisely calculate $\boldsymbol{\Delta\mu}^c$ and $\boldsymbol{\Sigma}_t^c$. The results are shown in Fig. \ref{Fig_iter_mem}. We observe that the mean and covariance estimations by our memory module manner better approximate the ideal estimation. Based on the results, we can infer that our manner better contributes to deriving more precise mean and covariance estimations.

\begin{figure}[htbp]
  \vspace{-2mm}
  \setlength{\abovecaptionskip}{0.cm}
  \setlength{\belowcaptionskip}{0.cm}
  \centering
  \includegraphics[width=0.37\textwidth]{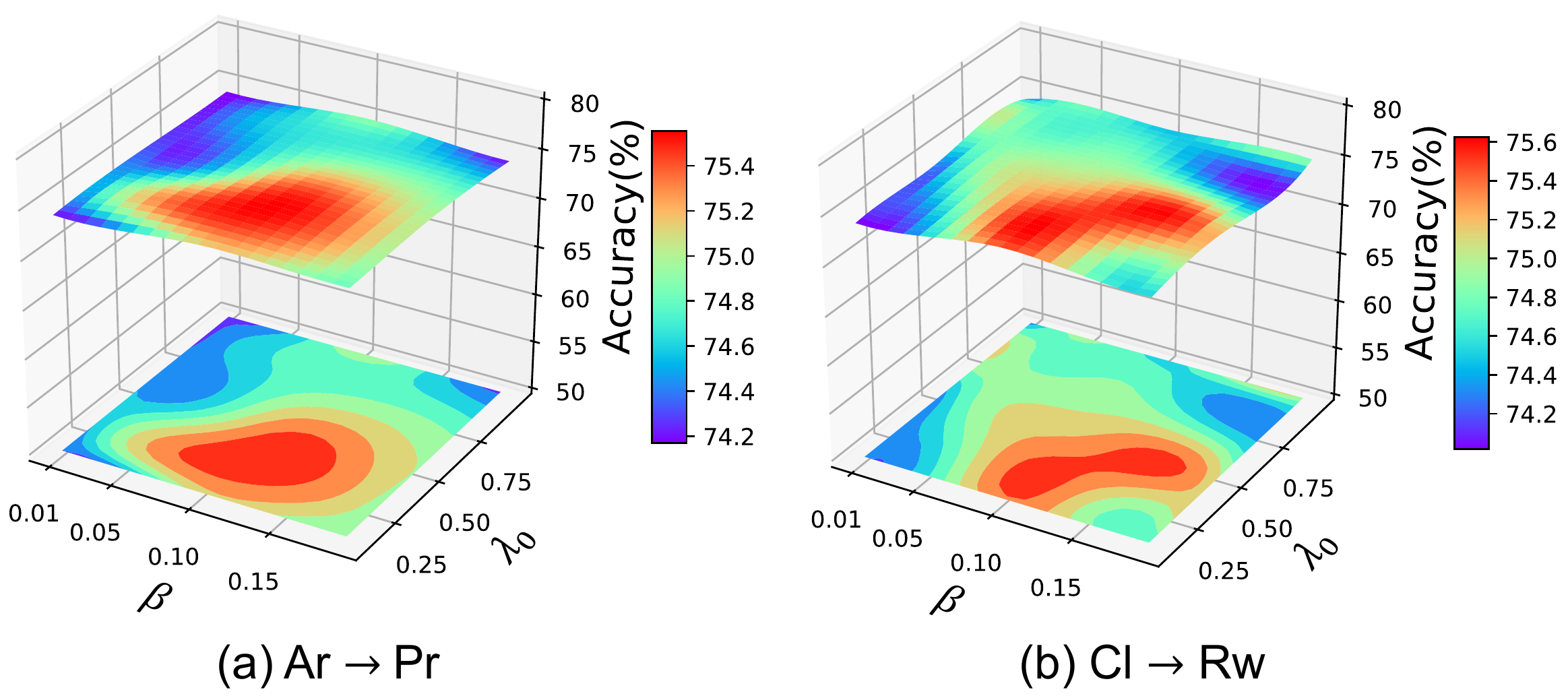}
  \caption{Hyper-parameter sensitivity analysis of TSA on tasks Ar$\rightarrow$Pr and Cl$\rightarrow$Rw (Office-Home) based on CDAN.}
  \label{Fig_sensitivity}
  \vspace{-2mm}
\end{figure}

\textbf{Hyper-parameter Sensitivity}. Hyper-parameter $\lambda_0$ controls the transformation strength and $\beta$ balances the two losses in TSA. To check the parameter sensitivity of TSA, we conduct experiments on two random tasks Ar$\rightarrow$Pr and Cl$\rightarrow$Rw by varying $\lambda_0\in\{0.1, 0.25, 0.5, 0.75, 1.0\}$ and $\beta\in\{0.01, 0.05, 0.1, 0.15, 0.2\}$. Fig. \ref{Fig_sensitivity} shows that TSA is not that sensitive to $\lambda_0$ and $\beta$, and can achieve competitive results under a wide range of hyper-parameter values.

\section{Conclusion}
This paper presents a transferable semantic augmentation (TSA) approach to ameliorate the adaptation ability of the classifier by optimizing a novel transferable loss over the implicitly augmented source distribution, introducing negligible computational costs. TSA is applicable to various DA methods and can yield significant improvements. Comprehensive experiments on several cross-domain datasets have demonstrated the efficacy and versatility of TSA.

{\small
\bibliographystyle{ieee_fullname}
\bibliography{Reference_CVPR2021}
}

\clearpage

\section*{Supplementary Materials}

\appendix

This supplementary materials provide the training process of Transferable Semantic Augmentation (TSA), the analysis of memory cost and performance gain, the algorithm for reverse mapping, extra visualization of the augmentation, the necessity stress test of target data and the visualization of learned features.   

\section{Training of TSA}

In summary, the proposed Transferable Semantic Augmentation (TSA) can be simply implemented and optimized in an end-to-end deep learning framework through the overall objective function:
\begin{small}
    \begin{align}
        \label{Eq:appendix_loss_all}
        \mathcal{L}_{TSA}&={\mathcal{L}}_{\infty}+\beta\mathcal{L}_{MI} \nonumber \notag\\
        &= - \frac{1}{n_{s}}\sum_{i=1}^{n_{s}}\log\frac{e^{{Z}^{y_{si}}_{si}}}{\sum_{c=1}^{C}e^{{Z}^{c}_{si}}} \notag\\
        &+ \beta(\sum_{c=1}^{C}\hat{P}^{c} \log \hat{P}^{c} - \frac{1}{n_{t}}\sum_{j=1}^{n_{t}}\sum_{c=1}^{C}{P_{tj}^{c}} \log P_{tj}^{c}),
    \end{align}
\end{small}where ${\mathcal{L}}_{\infty}$ is our proposed transferable loss and $\mathcal{L}_{MI}$ is the mutual information maximization loss.

The training process of TSA for domain adaptation is presented in Algorithm \ref{alg:1}.

\begin{algorithm}
    \renewcommand{\algorithmicrequire}{\textbf{Input:}}
    \renewcommand{\algorithmicensure}{\textbf{Output:}}
    \caption{Transferable Semantic Augmentation (TSA)}
    \label{alg:1}
    \begin{algorithmic}[1]
        \REQUIRE Labeled source domain $\mathcal{S}=\{(\x_{si}, y_{si})\}_{i=1}^{n_{s}}$, unlabeled target domain $\mathcal{T}=\{\x_{tj}\}_{j=1}^{n_{t}}$; maximum iteration $T$ and batch size $B$; hyper-parameters: $\lambda_{0}$ and $\beta$.
        \ENSURE Parameters of the final model: $\boldsymbol{\Theta}_{F}$, $\mathbf{W}$ and $\boldsymbol{b}$.
        \STATE Initialize model parameters $\boldsymbol{\Theta}_{F}$, $\mathbf{W}$ and $\boldsymbol{b}$; and initialize memory module $\mathbb {M}$ with $\mathcal{S}$ and $\mathcal{T}$.
        \FOR{$t=1$ to $T$}
            \STATE $\lambda = (t/T)\times{\lambda_{0}}$ \COMMENT{$\lambda$ controls augmentation strength}
            \STATE Sample $\{(\x_{si}, y_{si})\}_{i=1}^{B}$ and $\{(\x_{tj})\}_{j=1}^{B}$ from $\mathcal{S}$ and $\mathcal{T}$, respectively.
            \STATE Obtain deep features $\{\mathbf{f}_{si}\}_{i=1}^{B}, \{\mathbf{f}_{tj}\}_{j=1}^{B}$ and logit outputs $\{\hat{\boldsymbol{y}}_{si}\}_{i=1}^{B}, \{\hat{\boldsymbol{y}}_{tj}\}_{j=1}^{B}$ for source and target samples, respectively.
            \STATE Compute probabilistic outputs of target samples: $\{\boldsymbol{P}_{tj} = softmax(\hat{\boldsymbol{y}}_{tj})\}_{j=1}^{B}$ and generate target pseudo labels: $\{y^\prime_{tj}=\mathop{\arg\max}_{c}{P^c_{tj}}\}_{j=1}^B$ .
            \STATE Update memory module $\mathbb {M}$ with features and pseudo labels in current batch $t$.
            \FOR {each class $c$}
                \STATE Estimate features means $\boldsymbol{\mu}_{s}^{c}$ and $\boldsymbol{\mu}_{t}^{c}$ according to memory module $\mathbb {M}$.
                \STATE Estimate inter-domain mean difference $\boldsymbol{\Delta\mu}^{c} = \boldsymbol{\mu}_{t}^{c} - \boldsymbol{\mu}_{s}^{c}$ .
                \STATE Estimate the target intra-class covariance matrix $\boldsymbol{\Sigma}^{c}_{t}$ according to memory module $\mathbb {M}$.
            \ENDFOR
            \STATE Update $\boldsymbol{\Theta}_{F}$, $\mathbf{W}$ and $\boldsymbol{b}$ by minimizing the loss $\mathcal{L}_{TSA}$ in Eq \eqref{Eq:appendix_loss_all} with stochastic gradient descent (SGD).
        \ENDFOR
    \end{algorithmic}
\end{algorithm}

\begin{figure*}[t]
  \begin{center}
  \setlength{\abovecaptionskip}{0.cm}
  \setlength{\belowcaptionskip}{0.cm}
  \includegraphics[width=1\textwidth]{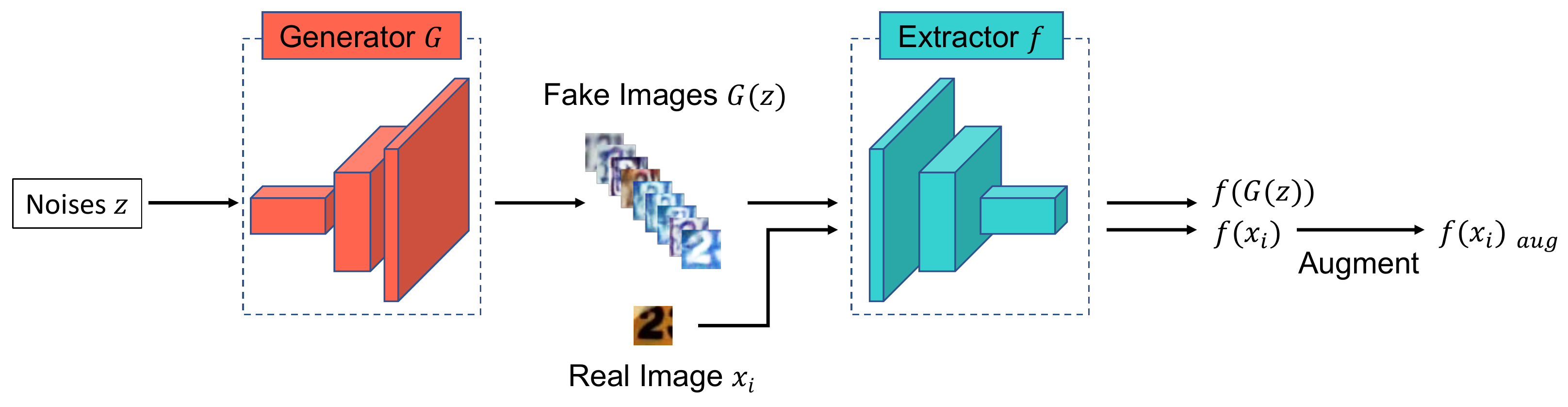}
  \caption{Outline of the reverse mapping algorithm. Generator $G$ is to generate fake images visually similar to source domain, while extractor $F$ is to extract the features of fake and source images. We use the two modules to search images in the pixel space corresponding to the augmented features in the deep feature space.}
  \label{Fig_visual_algorithm}
  \end{center}
  \vspace{-5mm}
\end{figure*}

\section{Memory Cost and Performance Gain}

In Table \ref{tab:memory}, we analyze the GPU memory cost and performance gain of our memory module $\mathbb{M}$. TSA (w/ iterative) denotes that we employ the iterative manner in \cite{ISDA} to estimate the mean and covariance, while TSA (w/ memory) denotes that we estimate the mean and covariance according to our memory module $\mathbb{M}$. In our experimental settings, the feature dimensions are set as 256 for Office-31. Compared to ResNet-50, TSA (w/ memory) obtains a large improvement of 13.2\% with negligible extra 0.3GB GPU memory cost, surpassing TSA (w/ iterative) by 2.8\%. This is mainly due to the fact that the iterative manner will bias the estimation of expected mean and covariance, due to its accumulation property. Besides, it is noteworthy that the memory module is not required in the inference phase.

\begin{table}[htbp]\small
  \centering
  \setlength{\abovecaptionskip}{0.cm}
  \setlength{\belowcaptionskip}{0.cm}
  \caption{GPU memory cost on Office-31 with batch-size 32.}
  \setlength{\tabcolsep}{0.4mm}{
    \begin{tabular}{|l|c|c|c|}
    \hline
    Method & GPU Memory (GB) & Accuracy (\%) & Gain (\%) \\
    \hline
    \hline
    ResNet-50 & 7.2 &  76.1 & - \\
    \hline
    TSA (w/ iterative) & 7.3 & 86.5 & 10.4 $\uparrow$ \\
    \hline
    TSA (w/ memory) & 7.5 & 89.3 & 13.2 $\uparrow$ \\
    \hline
    \end{tabular}}
  \label{tab:memory}
\end{table}

\textbf{Iterative Estimation Manner.} Here, we elaborate the iterative manner of updating mean and covariance. To estimate the covariance matrix of features, ISDA \cite{ISDA} proposes an iterative manner by integrating covariances of batches from first to current batch. For class $c$, the iterative estimation manner is formulated as:
\begin{align}
  \eta_{c}^i & = \frac{B_{c}^i}{N_{c}^{(i-1)}+B_{c}^i},\\
  \boldsymbol{\mu}_{c}^{(i)}&=(1-\eta_{c}^{i})\boldsymbol{\mu}_{c}^{(i-1)}+\eta_{c}^{i}\boldsymbol{\mu^\prime}_{c}^{i},\\
  \boldsymbol{\Sigma}_{c}^{(i)}&=(1-\eta_{c}^i)\boldsymbol{\Sigma}_{c}^{(i-1)}+\eta_{c}^i\boldsymbol{\Sigma^\prime}_{c}^i\notag\\
  &+\eta_{c}^i(1-\eta_{c}^i)(\boldsymbol{\mu}_{c}^{{(i-1)}}-\boldsymbol{\mu^\prime}_{c}^{i})(\boldsymbol{\mu}_{c}^{(i-1)}-\boldsymbol{\mu^\prime}_{c}^{i})^{\top},
\end{align}where $N_{c}^{(i-1)}$ is the total number of training samples for class $c$ in all previous $(i-1)$ batches, and $B_{c}^i$ is the number of training samples belonging to class $c$ in current batch $i$. $\eta_{c}^i$ is the proportion of samples of class $c$ in current batch $i$ to samples of class $c$ in all previous $(i-1)$ batches. $\boldsymbol{\mu}_{c}^{(i-1)}$ and $\boldsymbol{\mu^\prime}_{c}^{i}$ are the averages of features for class $c$ in all previous $(i-1)$ batches and current batch $i$, respectively. $\boldsymbol{\Sigma}_{c}^{(i-1)}$ and $\boldsymbol{\Sigma^\prime}_{c}^{i}$ are the covariances of features for class $c$ in all previous $(i-1)$ batches and current batch $i$, respectively.

However, the weight distribution of network in early training stage will vastly differ from that in latter training stage. Thus, due to its accumulation property, such iterative manner might cause out-of-date features to bias the estimation of expected covariance matrix. By contrast, our memory module will discard the out-of-date batch and replace with the latest batch, conductive to more accurate estimations of mean and covariance.

\begin{figure*}[htbp]
  \centering
  \setlength{\abovecaptionskip}{0.2cm}
  \setlength{\belowcaptionskip}{0.cm}
  \includegraphics[width=1.0\textwidth]{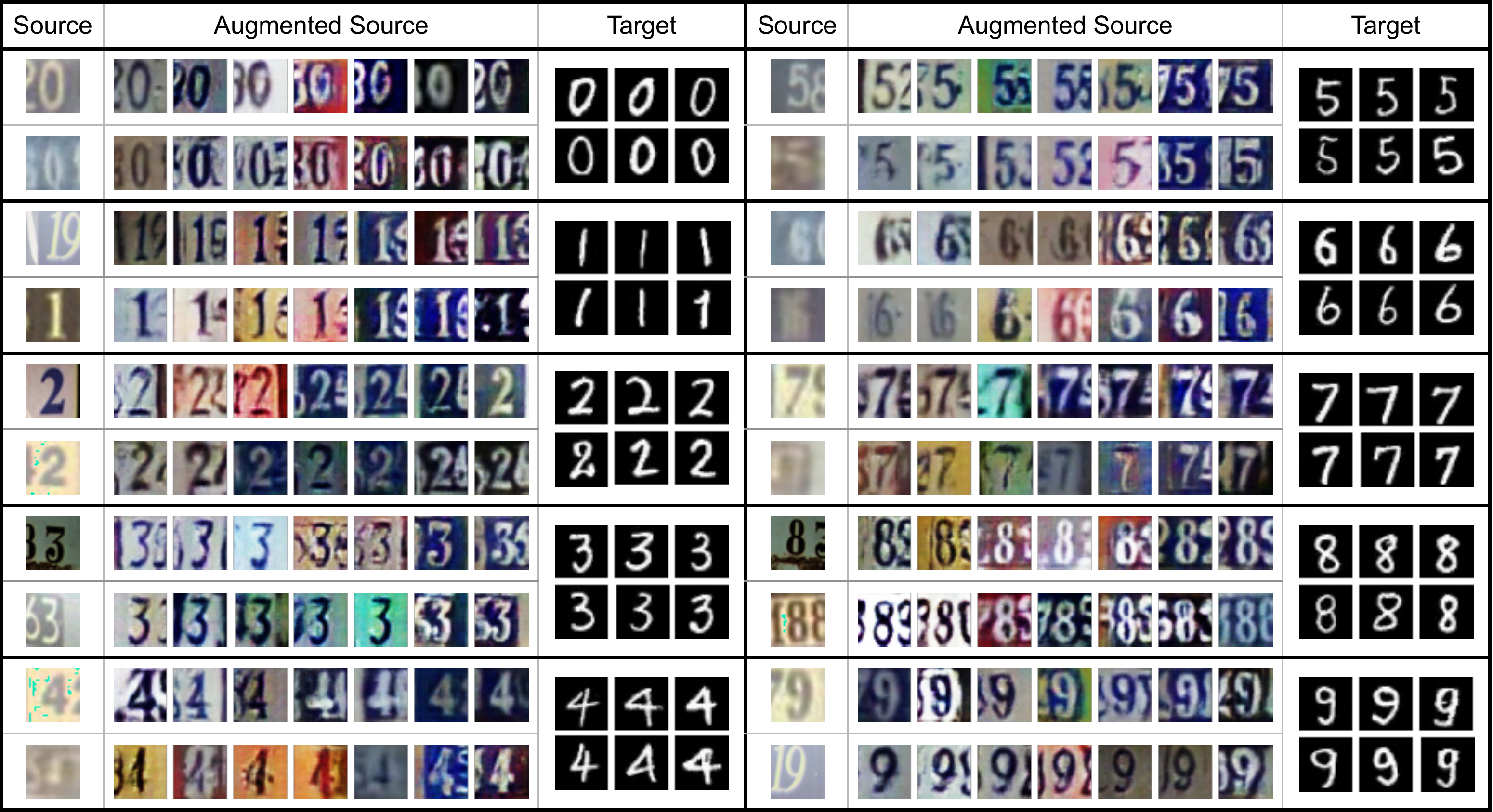}
  \caption{Extra visualizations of semantically augmented images for task SVHN$\rightarrow$MNIST. ``Source'' and ``Augmented Source'' columns present the images from SVHN and their corresponding augmented images synthesized by TSA, respectively. ``Target'' column provides several images representing corresponding classes from MNIST.}
  \label{Fig_extra_samples}
  \vspace{-3mm}
\end{figure*}

\section{Reverse Mapping Algorithm}

To intuitively display how TSA generates the meaningful semantic transformations of source images, inspired by \cite{ISDA}, we design a reverse mapping algorithm to generate images in the pixel-level space corresponding to the augmented features in the deep feature space.

The outline of the reverse mapping algorithm is shown in Fig. \ref{Fig_visual_algorithm}. There are two modules involved in this algorithm: 1) GANs-based generator $G$ and 2) CNNs-based feature extractor $F$. Here, the generator $G$ is pre-trained on a source dataset to generate fake images that are close to source domain, while feature extractor $F$ is pre-trained on an adaptation task to extract the features of fake and source images. Taking the adaptation task SVHN$\rightarrow$MNIST as an example, $G$ is pre-trained on dataset SVHN, and $F$ is pre-trained on task SVHN$\rightarrow$MNIST. After the two modules are well trained separately, we freeze them during following phases. 

Formally, let $\z\in\mathbb{R}^{d}$ denote the random noise variable and $\x_{s}$ denote the source image randomly sampled from a source domain, such as SVHN. At first, noise $\z$ is fed into generator $G$ to synthesize fake image ${G(\z)}$. Then, fake image $G(\z)$ and source image $\x_{s}$ are input to the feature extractor $F$ to obtain deep features $F(G(\z))$ and $F(\x_{s})$, respectively. For achieving reliable visualization results, we first find the noise $\z'$ corresponding to source image $\x_{s}$ by solving the following optimization problem:
\begin{small}
\begin{align}
  \label{eq:reverse1}
  \z' = \arg\mathop{\min}_{\z}\|F(G(\z))-F(\x_{s}) \|_{2}^{2} + \alpha\|G(\z)-\x_{s}\|_{2}^{2},
\end{align}
\end{small}where $\alpha$ is a trade-off parameter to balance the contributions of two parts. We use the mean square error to constrain noise $\z$ in both feature and pixel levels to obtain a reliable initial point $\z'$ for the following procedure.

After yielding the initial point $\z'$, we augment the original source feature $F(\x_{s})$ with TSA, forming the augmented feature $F_{aug}(\x_{s})$. Then we initialize $\z$ with $\z'$ to continue to search the optimal noise $\z^{*}$ corresponding to the augmented feature $F_{aug}(\x_{s})$:
\begin{align}
  \label{eq:reverse2}
  \z^{*} = \arg\mathop{\min}_{\z}\|F(G(\z))-F_{aug}(\x_{s})\|_{2}^{2}.
\end{align}When Eq \eqref{eq:reverse2} is optimized, the desired $\z^{*}$ can be obtained. Consequently, the image $G(\z^{*})$ generated by $G$ should be the appropriate visualization for the corresponding augmented source feature.

\textbf{Implemented Details}. We adopt the structure of WGAN-GP \cite{WGAN-GP} for the generator $G$. The architecture of feature extractor $F$ is based on the chosen task, e.g., we use the feature extractor of task SVHN$\rightarrow$ MNIST as \cite{MCD,JADA}. We also adopt SGD optimizer with momentum $0.9$ to optimize the reverse mapping algorithm. The dimension $d$ of noise variable is set to be $128$.

\textbf{Extra Visualization Results of Augmentation}. We adopt the aforementioned reverse mapping algorithm to produce more visualization results for augmented source features, which are shown in Fig. \ref{Fig_extra_samples}. The results further prove that TSA is indeed able to generate diverse, meaningful and even target-specific augmented features, which will facilitate adapting the classifier from source to target domain successfully.

\section{Necessity Stress Test of Target Data}

For reducing the training burden in practice, we test how much target data is necessary to achieve desirable performance of TSA by varying the amount of target data participated in the training. Specifically, for each class, we randomly sample a proportion ($\rho$) of target data as training samples along with the original source dataset, and use the entire target dataset for evaluating. The results of ResNet-50+TSA on Office-31 under $\rho \in \{0\%, 20\%, 40\%, 60\%, 80\%, 100\%\}$ are shown in Table \ref{tab:Necessity stress tests}, where we can see that when target data increases, higher accuracies are obtained. This is because more accurate and comprehensive semantics can be captured with more target data. Besides, the performances of $\rho=80\%$ and $\rho=100\%$ are comparable, which motivates us to reduce target data in training to save computation and memory cost for large-scale target datasets.

\begin{table}[htbp]
  \footnotesize
  \setlength{\abovecaptionskip}{0.cm}
  \setlength{\belowcaptionskip}{0.cm}
  \centering
  \caption{Necessity stress test of target data on Office-31.}
    \setlength{\tabcolsep}{0.6mm}{
    \begin{tabular}{|c|c|cccccc|c|}
    \hline
    Method & $\rho$ & A$\rightarrow$W & D$\rightarrow$W & W$\rightarrow$D & A$\rightarrow$D & D$\rightarrow$A & W$\rightarrow$A & Avg \\
    \hline
    \multirow{6}{*} {\makecell*[c]{ResNet-50\\+\\TSA}} & 0\% & 68.4 & 96.7 & 99.3 & 68.9 & 62.5& 60.7 & 76.1 \\
    \cline{2-9} & 20\% & 82.4 & 97.4 & 99.5 & 81.5 & 66.4 & 64.8 & 82.0 \\
    \cline{2-9} & 40\% & 90.2 & 98.4 & 99.8 & 88.5 & 70.5 & 69.7 & 86.2 \\
    \cline{2-9} & 60\% & 93.2 & 98.8 & \textbf{100.0} & 91.7 & 73.4 & 73.0 & 88.4 \\
    \cline{2-9} & 80\% & 94.3 & 98.8 & \textbf{100.0} & 92.4 & 74.8 & 74.2 & 89.1 \\
    \cline{2-9} & 100\% & \textbf{94.8} & \textbf{99.1} & \textbf{100.0} & \textbf{92.6} & \textbf{74.9} & \textbf{74.4} & \textbf{89.3} \\
  \hline  
  \end{tabular}}%
  \label{tab:Necessity stress tests}
  \vspace{-3mm}
\end{table}%

\begin{figure}[htbp]
  \centering
  \setlength{\abovecaptionskip}{0.1cm}
  \setlength{\belowcaptionskip}{0.cm}
  \includegraphics[width=0.47\textwidth]{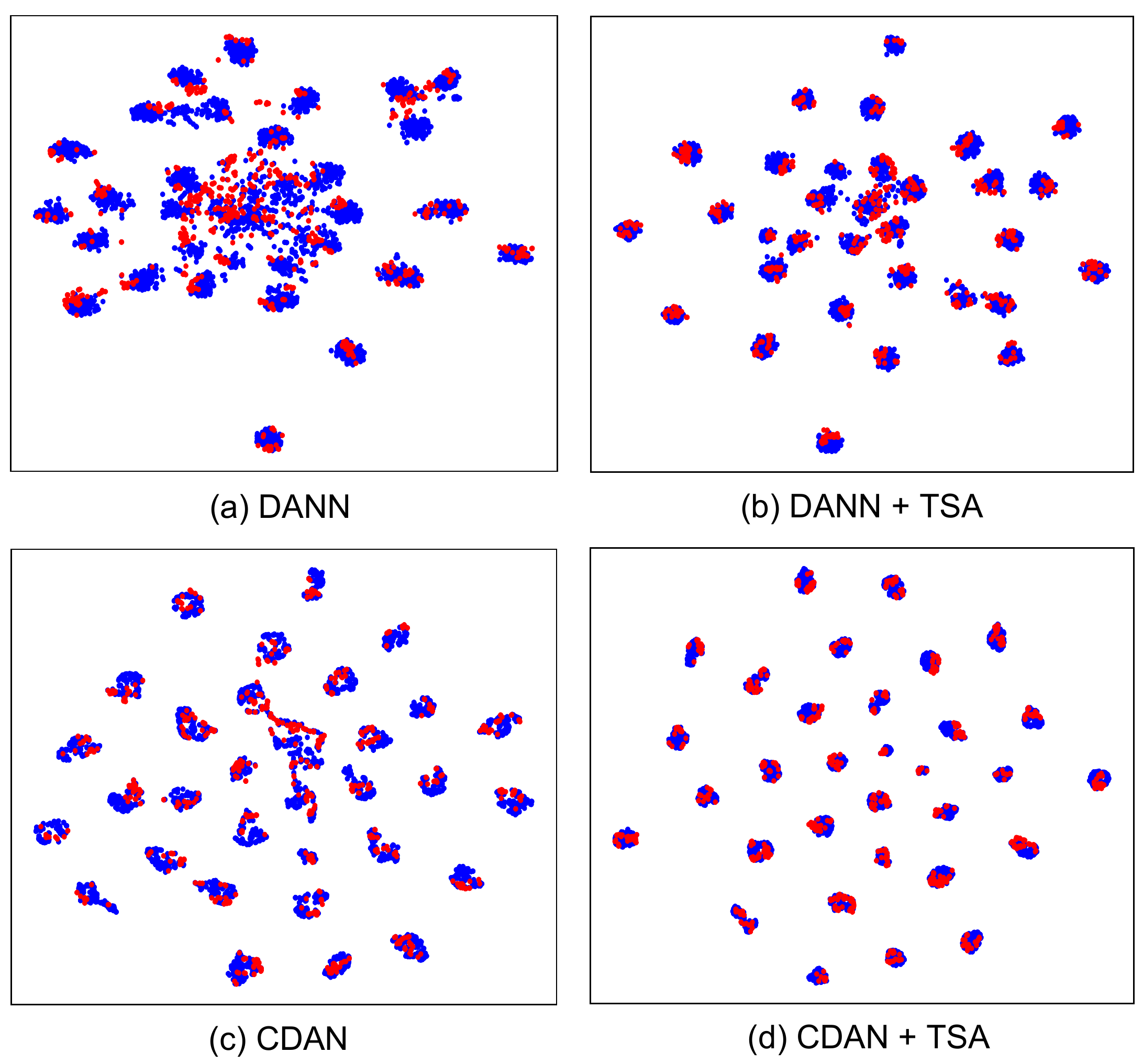}
  \caption{Visualization of the features learned by DANN, DANN+TSA, CDAN and CDAN+TSA on task A$\rightarrow$W (Office-31). Blue and red dots stand for the source features and the target features, respectively.}
  \label{Fig_tsne}
  \vspace{-3mm}
\end{figure}

\section{Visualization of Features}
Though TSA strives to adapt classifiers from source domain to target domain, we surprisingly notice that TSA can also empower DA methods to learn more transferable feature representations. We adopt t-SNE tool \cite{tsne} to visualize the features learned by DANN \cite{DANN}, DANN+TSA, CDAN \cite{CDAN}, and CDAN+TSA on task A${\rightarrow}$W (Office-31) in Fig. \ref{Fig_tsne}. From results we observe that DANN and CDAN cannot align both domains perfectly, due to the mismatching of several classes. After applying TSA, the representations are more indistinguishable between two domains and the class boundaries are sharper, validating that TSA facilitates learning more transferable and discriminative features besides the effective classifier adaptation.

\end{document}